\documentclass{article}

\usepackage{microtype}
\usepackage{graphicx}
\usepackage{subcaption}
\usepackage{booktabs} 
\usepackage{bbm}
\usepackage[table]{xcolor}
\usepackage{hyperref}

\usepackage[preprint]{icml2026}

\usepackage{amsmath}
\usepackage{amssymb}
\usepackage{mathtools}
\usepackage{amsthm}
\usepackage{tikz}
\usepackage{pgfplots}
\usepackage{float}
\usepgfplotslibrary{groupplots}
\pgfplotsset{compat=1.18}

\usepackage[capitalize,noabbrev]{cleveref}

\theoremstyle{plain}

\theoremstyle{definition}

\theoremstyle{remark}

\usepackage{multirow}

\usepackage[textsize=tiny]{todonotes}

\icmltitlerunning{Submission and Formatting Instructions for ICML 2026}

\begin{document}

\twocolumn[
  \icmltitle{Large Language Models Decide Early and Explain Later}

    








  \icmlsetsymbol{equal}{*}

  \begin{icmlauthorlist}
    \icmlauthor{Ayan Datta}{iiit}
    \icmlauthor{Zhixue Zhao}{sheffield}
    \icmlauthor{Bhuvanesh Verma}{goethe}
    \icmlauthor{Radhika Mamidi}{iiit}
    \icmlauthor{Mounika Marreddy}{goethe}
    \icmlauthor{Alexander Mehler}{goethe}
  \end{icmlauthorlist}

  \icmlaffiliation{iiit}{IIIT Hyderabad}
  \icmlaffiliation{sheffield}{University of Sheffield, UK}
  \icmlaffiliation{goethe}{Goethe University, Frankfurt am Main, Germany}

  \icmlcorrespondingauthor{Ayan Datta}{advin4603@gmail.com}

  \icmlkeywords{Machine Learning, ICML}

  \vskip 0.3in
]



\printAffiliationsAndNotice{}  


\begin{abstract}
Large Language Models often achieve strong performance by generating long intermediate chain-of-thought reasoning. However, it remains unclear when a model’s final answer is actually determined during generation. If the answer is already fixed at an intermediate stage, subsequent reasoning tokens may constitute post-decision explanation, increasing inference cost and latency without improving correctness. We study the evolution of predicted answers over reasoning steps using forced answer completion, which elicits the model’s intermediate predictions at partial reasoning prefixes. Focusing on Qwen3-4B and averaging results across all datasets considered, we find that predicted answers change in only \textbf{32\%} of queries. Moreover, once the final answer switch occurs, the model generates an average of \textbf{760 additional reasoning tokens per query}, accounting for a substantial fraction of the total reasoning budget. Motivated by these findings, we investigate early stopping strategies that halt generation once the answer has stabilized. We show that simple heuristics, including probe-based stopping, can reduce reasoning token usage by \textbf{500 tokens per query} while incurring only a \textbf{2\% drop in accuracy}. Together, our results indicate that a large portion of chain-of-thought generation is redundant and can be reduced with minimal impact on performance.
\end{abstract}


\vspace{-0.50cm}

\begin{figure}[t]
    \centering
    \includegraphics[width=.88\linewidth]{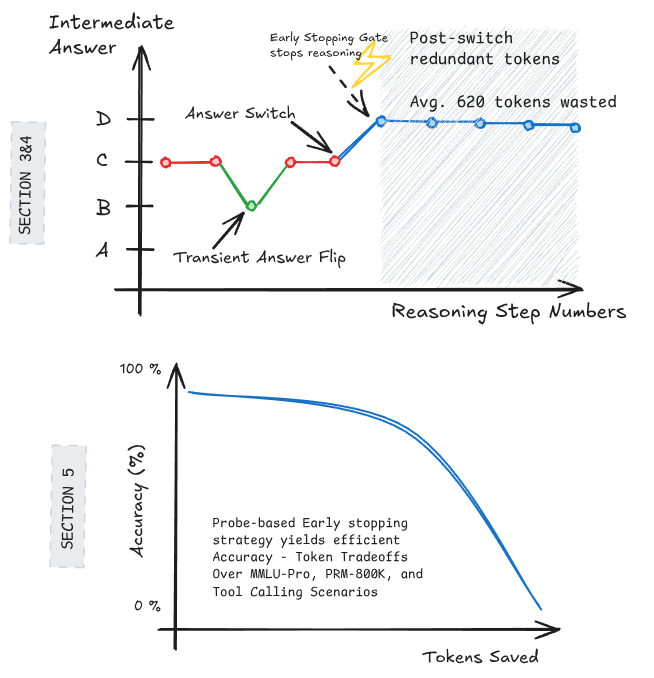}
    \caption{Top: illustration of an obtained answer trajectory for a multiple-choice task. A brief transient flip (answer C → answer B → answer C) reflects local instability, whereas the later sustained transition to D constitutes a genuine answer switch. We study this in Section~\ref{sec:results}. Bottom: Early-stopping to save token usages with (detailed figures are in Section~\ref{sec:early_stopping})}
    \label{fig:placeholder}
\end{figure}

\section{Introduction}
Large language models (LLMs) trained for explicit reasoning have achieved strong performance across a wide range of reasoning tasks, including mathematical problem solving, multiple choice question answering, and tool assisted decision making~\cite{wei2022chain, plaat2025multi, lewkowycz2022solving, kojima2022large}, by generating intermediate reasoning steps before producing a final answer. While these traces are typically treated as a monolithic process, the temporal relationship between individual reasoning steps and the model’s predicted answer remains poorly understood.
Despite their empirical success, the role of these intermediate reasoning tokens in the actual decision-making process of LLMs is unclear. While longer reasoning traces are often associated with improved accuracy, it is not evident whether each generated token faithfully contributes to updating the model’s internal belief about the correct answer. In practice, models may continue to generate coherent reasoning even after their beliefs have effectively stabilized. This raises the possibility that a significant portion of reasoning tokens are redundant to the reasoning and do not influence the final output.


Recent studies have questioned the faithfulness of chain-of-thought explanations. Prior work has shown that models can produce correct answers accompanied by incorrect or misleading rationales~\cite{wan2024evidence, wiegreffe2019attention}, and that reasoning traces can be altered or shortened with little impact on model accuracy~\cite{zhou2022least, wang2022self}. Together, these findings suggest that textual reasoning may not reliably reflect the underlying computations driving a model’s decisions. However, most existing analyses treat reasoning as a static object, focusing on the presence or absence of chains of thought, rather than examining how a model’s answer evolves during the reasoning process. On the other hand, the cost of verbose reasoning has become increasingly important in practice. Long reasoning traces increase inference latency, computational cost, and energy consumption, reduce output conciseness, and can complicate deployment in tool-based or agentic systems~\cite{yao2022react, karpas2022mrkl}. Despite these concerns, there has been limited work investigating how models' predictions stabilize over the course of the reasoning process.

To address this, we use \emph{forced answer completion}, a framework that intervenes on partial reasoning prefixes to elicit the model’s predicted answer at intermediate steps. By applying this procedure throughout a reasoning trace, we obtain a \emph{dynamic answer trajectory} that records how the predicted answer changes over reasoning steps generated. This makes it possible to measure answer switches, transient answer flips, and the point at which the predicted answer reaches its final decision.

Using answer trajectories, we empirically characterize the temporal dynamics of predicted answers throughout the model’s reasoning process across multiple task settings.\footnote{Unless stated otherwise, all statistics reported in this section are computed using Qwen3-4B and averaged over all datasets considered.} We find that predicted answers change in only \textbf{32\%} of queries on Qwen3-4B, indicating that in the majority of cases the model’s answer is fixed from the earliest stages of generation. When answer changes do occur, they typically happen early in the reasoning trace. Despite this early stabilization, the model continues to generate substantial additional reasoning, producing on average \textbf{760 tokens after the final answer switch}, and up to \textbf{1000 tokens} in some cases. This reveals a systematic temporal gap between when a model’s decision is made and when its explanation is produced.

\section{Related Work}
\paragraph{Faithfulness and Necessity of Chain-of-Thought.}
Chain-of-thought (CoT) prompting improves performance on many reasoning tasks by encouraging models to generate intermediate steps before giving a final answer~\cite{wei2022chain, kojima2022large}. However, recent work questions whether these reasoning steps truly reflect how models reach their final answers. Prior studies show that models can produce correct answers even when their explanations are incorrect, misleading, or added after the fact~\cite{wan2024evidence}. Other work finds that reasoning traces can be edited, rewritten, or significantly shortened with little effect on accuracy, suggesting that not all generated reasoning is necessary for the final prediction~\cite{zhou2022least, wang2022self}. Together, these results suggest that many reasoning tokens may not directly influence the model’s decision.

Most existing analyses treat CoT as a static output and focus on whether a final explanation is present, well-formed, or faithful. In contrast, our work studies reasoning as a dynamic process, examining how a model’s predicted answer changes as reasoning tokens are generated.

\paragraph{Efficient Reasoning and Token Economy.}
As reasoning-enabled models become more widely deployed, the computational cost of long reasoning traces has emerged as a practical concern. Verbose reasoning increases inference latency, memory usage, and energy consumption, and can complicate deployment in tool-augmented or agentic systems~\cite{yao2022react, karpas2022mrkl}. Several recent works have therefore explored methods for reducing reasoning cost while preserving performance. These include approaches that compress~\cite{}, distill~\cite{}, or restructure reasoning traces~\cite{}, as well as methods that encourage models to produce shorter or more selective chains of thought~\cite{zhou2022least}. More recent studies have begun to explicitly analyze reasoning token wastage, showing that many generated reasoning tokens contribute little to final answer accuracy and may constitute post-decision explanations rather than necessary computation~\cite{bogdan2025thought}. Related work has also examined adaptive computation and efficiency-oriented decoding strategies that aim to balance reasoning depth against computational cost~\cite{lanham2023measuring}. However, these approaches typically rely on heuristic stopping criteria or global constraints on reasoning length, without directly measuring when a model’s answer has effectively stabilized.

\paragraph{Answer Dynamics and Early Stopping.}
A smaller line of work has investigated intermediate predictions during generation, often in the context of calibration~\cite{mei2025reasoning,mao2026confidence}, uncertainty estimation~\cite{mei2025reasoning,yoon2025reasoning,zhang2025cot,mao2026recurrent}, or self-consistency~\cite{wang2022self,liang2024internal}. These methods implicitly acknowledge that a model’s belief about the correct answer may evolve, but they do not directly characterize the temporal structure of answer changes within a single reasoning trace. Our work builds on these insights by explicitly tracing model predictions at intermediate reasoning steps. 

In summary, prior work has established that CoT reasoning is often unfaithful, unnecessarily verbose, and costly, but has largely analyzed reasoning as a static object or focused on global efficiency constraints. In contrast, our work offers a temporal, answer-centric analysis of reasoning, quantifying answer switches, post-decision token generation, and the potential gains from early stopping. 

\section{Methodology}

\subsection{Problem Setup and Notation}

We study tasks in which a language model is given an input context and produces a final answer accompanied by an explicit reasoning trace. The input context is denoted by $C$, which includes the conversation history, task prompt, and any additional information provided to the model. Let $R$ denote the reasoning trace generated by the model. We represent $R$ as a sequence $R = (R_1, R_2, \ldots, R_n)$, where each $R_i$ corresponds to a distinct unit of reasoning. In this work, we assume individual sentences as the basic units of reasoning for analysis purposes, consistent with prior work on natural language rationales and chain-of-though explanations~\cite{wei2022chain, lampinen2022can}. Recent reasoning-oriented models explicitly separate reasoning from the final answer. In particular, models inspired by~\citet{shao2024deepseekmath} employ special \texttt{<think>} tags to delimit reasoning content. Under this formulation, the final assistant output can be written as: $\texttt{<think>}R\texttt{</think>}A$, where $A$ denotes the final answer.

\subsection{Forced Answer Completion}
Inspired by~\citet{lanham2023measuringfaithfulnesschainofthoughtreasoning}, we analyze the evolution of model predictions over the course of a reasoning trace. Given an input context $C$ and a reasoning trace $R = (R_1, R_2, \ldots, R_n)$, we estimate the intermediate answer the model would predict after each step by explicitly intervening to halt further reasoning and elicit an immediate final answer. Concretely, after generating the partial reasoning trace $(R_1, \ldots, R_i)$, we append the end-of-thinking token \texttt{</think>} followed by an answer prefix $P$, yielding the prompt
\[
C\texttt{<think>}R_1 R_2 \ldots R_i\texttt{</think>}P.
\]
Sampling the model conditioned on this prompt produces an intermediate answer $A_i$. Repeating this procedure after each reasoning step yields an \emph{answer trajectory} $T = (A_0, A_1, \ldots, A_n)$, where $A_0$ denotes the answer generated without any reasoning steps and $A_i$ denotes the answer generated after step $R_i$. The answer prefix for each task is listed in Section~\ref{app:answer_prefix} in the Appendix.

\subsection{Measuring Decision Changes}
\label{sec:answer-trajectory-metrics}
Given an Answer Trajectory $T = (A_0, A_1, A_2, \ldots, A_n)$ induced by forced answer completion, we define a set of metrics that characterize the temporal behaviour of the model's predicted answer over the reasoning steps $R = (R_1, R_2, \ldots R_n)$.

\subsubsection{Final answer switch index.}
\label{sec:final_answer_switch}
We define the index of the last answer switch as the index after which the model's intermediate answer does not change. We calculate this as:
\begin{equation}\small
t^\star
\;=\;
\max \big\{\, i \in \{0,\ldots,n-1\} \;\big|\; A_i \neq A_{i+1} \,\big\},
\end{equation}
with the convention that $t^\star = -1$ if no such index exists (i.e., $A_0 = A_1 = \cdots = A_n$).

\subsubsection{Answer Switches}

We calculate the number of times the model switches its answer from one to another. We do so using the following:

\begin{equation}\small
    \text{AnswerSwitches}(T) = \sum_{i=0}^{n-1}[A_i \neq A_{i+1}]
\end{equation}

Where $[\cdot]$ denotes the Iverson bracket, which evaluates to $1$ if the condition is true, else $0$.

\subsubsection{Transient Answer Flips}

We also detect cases where the predicted answer briefly switches to an alternative, e.g., for one more step, and then switches back. We refer to these locally inconsistent trajectories \emph{Transient Answer Flips (TAFs)}. That is, a transient answer flip is a short-lived deviation from a dominant answer, lasting a small number of reasoning steps ($k$), after which the model returns to the original answer. Given an Answer Trajectory $T = (A_0, A_1, A_2, \ldots, A_n)$, we calculate the number of TAFs as follows:
\begin{equation}\small
\small
\text{TAFs}(T)
=
\sum_{i=0}^{n-k-1}
\left[
\begin{aligned}
&\exists \,\ell \le k \;\text{s.t.}\; A_i = A_{i+\ell+1} \\
&\land \;\forall j \in \{1,\dots,\ell\},\; A_i \neq A_{i+j}
\end{aligned}
\right]
\end{equation}
Where $k$ is the max length of TAFs we want to detect. We set $k = 3$ as a reasonable default.

\subsection{Answer Trajectory Denoising}
A high frequency of TAFs indicates that $A_i \neq A_{i+\ell}$ for a small number of further reasoning steps $\ell$, despite $A_i = A_{i+\ell+1}$. That is, the model briefly deviates from an answer and then quickly returns to its previous value. Such short-lived reversals are unlikely to correspond to changes in the model’s final decision, since they do not persist long enough to affect the eventual output. Instead, they typically reflect local instability in the intermediate reasoning trace, arising from transient token-level fluctuations or sensitivity to nearby context, rather than deliberate revisions of the underlying conclusion.

Motivated by this observation, we aim to filter out these short-lived deviations in the answer trajectory and retain only answer changes that are sustained over multiple consecutive reasoning steps. To this end, we apply a simple causal smoothing procedure that suppresses brief oscillations while preserving persistent answer switches. This denoising removes excess answer switches that do not meaningfully correspond to decision changes, but instead arise from momentary answer flips. As a result, the estimated final answer switch becomes less sensitive to noisy local fluctuations, revealing that even more post-decision reasoning tokens can be saved once such spurious flips are discounted.

\paragraph{Hold-for-$k$ Smoothing}
Given an answer trajectory $T = (A_0, A_1, A_2, \ldots, A_n)$, we define a causal smoothing operator, referred to as \emph{hold-for-$k$}. The operator enforces persistence by requiring an answer to remain unchanged for $k$ consecutive reasoning steps before it is accepted as a genuine answer switch. Intuitively, this prevents brief, noisy answer flips from being treated as meaningful changes.

Formally, the smoothed answer trajectory $\tilde{T}^{(k)} = (\tilde{A}^{(k)}_1, \ldots, \tilde{A}^{(k)}_n)$ is defined recursively as:
\begin{equation}\small
\tilde{A}^{(k)}_i =
\begin{cases}
A_i, & \text{if } A_{i-j} = A_i \;\; \forall j \in \{0, \ldots, k-1\}, \\
\tilde{A}^{(k)}_{i-1}, & \text{otherwise}.
\end{cases}
\end{equation}

This construction is causal, as the smoothed answer at step $i$ depends only on the current and previous reasoning steps and does not rely on future information. This property is essential for online or streaming settings, where answer trajectories are observed incrementally as the model generates its reasoning. In such settings, non-causal smoothing would require access to future steps and therefore cannot be applied in real time. By enforcing causality, hold-for-$k$ smoothing enables on-the-fly monitoring of answer trajectories, allowing sustained answer switches to be detected as they emerge during generation rather than retrospectively.

\subsection{Tokens After Final Answer Switch}
 
Given an answer trajectory $T = (A_0, \ldots, A_n)$, we measure the number of tokens generated after $t^\star$.
Let $T_i$ denote the cumulative number of generated reasoning tokens up to step $i$, with $T_0 = 0$ and $T_n = T_{\mathrm{total}}$.
We define:
$T_{\mathrm{after}}
\;=\;
T_{\mathrm{total}} - T_{t^\star+1}.
$
This quantity captures the amount of generated tokens that would be saved by halting reasoning immediately after the final answer switch, without altering the model’s predicted answer.

\subsection{Datasets}

We evaluated a total of four tasks. For each task, we report per-example trajectories $(A_0,\dots,A_n)$ under forced completion and apply identical evaluation and monitoring procedures. 
\vspace{-0.3cm}
\paragraph{Multiple-choice Question Answering (MCQ)} MCQ is commonly used for reasoning evaluation, where each example consists of a question and a small fixed set of candidate options, typically \(\{A,B,C,D\}\). We use the datasets from~\cite{wang2024mmluprorobustchallengingmultitask}, sampling the correct answer choice and three other distractor choices. For these datasets, we treat the answer space as the finite set of provided options. At each reasoning step $i$, we record the option with the highest probability under forced answer completion and use this option as the per-step label $A_i$.
\vspace{-0.3cm}
\paragraph{Numeric-answer} We use the dataset from~\cite{lightman2023lets}, where correct responses are predominantly numeric and typically represented by a single token (e.g., an integer or short numeric constant). For these tasks, we record the highest-probability token under forced completion at each generation step, yielding an answer trajectory $A_0,\dots,A_n$.
\vspace{-0.3cm}
\paragraph{Search-query}
We evaluate on a closed-book question-answering dataset from~\cite{monteiro2024repliqaquestionansweringdatasetbenchmarking} augmented with an explicit search-tool interface. Each example consists of a question paired with a collection of background documents. During generation, the model is prompted to invoke a search tool by producing an appropriate search query as a tool argument.

For analysis, we treat the generated search query itself as the answer. When forced answer completion causes the model to emit a search query, we extract the query text and compare it to the query produced under full generation using the answer label mapping rules. We do not execute the search; instead, we compare the generated queries directly and use query similarity as a proxy for retrieval equivalence.

During forced answer completion, we allow the model to fully generate the search query using the same sampling settings as used for the reasoning trace. To compare two queries obtained at reasoning steps $i$ and $j$, denoted $A_i$ and $A_j$, we embed each query using a fixed query encoder to obtain vector representations $\mathbf{a}_i$ and $\mathbf{a}_j$. We consider the two queries to be equivalent if their cosine similarity exceeds a fixed threshold.
\begin{equation}\small
A_i = A_j
\;\;\iff\;\;
\frac{\mathbf{a}_i^\top \mathbf{a}_j}{\|\mathbf{a}_i\| \, \|\mathbf{a}_j\|}
\;\ge\;
\gamma,
\end{equation}
where $\gamma = 0.9$ in all experiments.
\vspace{-0.3cm}
\paragraph{Tool-selection} Tool selection is a fundamental challenge in agent-based systems. We evaluate the reasoning capability of LLMs using tasks in which the model is given a fixed set of tools and must select exactly one tool as part of its response~\cite{mcp_tool_call_eval_test}. At each reasoning step, we extract the tool identifier produced by the model when prompted to emit a tool call. The selected tool is recorded as the per-step label \(A_i\), and the set of available tools defines the answer space for these datasets.

\subsection{Models}
We consider the Qwen3 family of language reasoning models. From Qwen3 we use the dense variants Qwen3-4B, Qwen3-8B and the MoE variant Qwen3-30B-A3B~\cite{yang2025qwen3technicalreport}. We use the query encoder from~\cite{reimers2019allMiniLML6v2} for comparing search queries.
For all models we use the recommended generation settings unless stated otherwise. Specifically, we set temperature \(=0.6\) for sampling during both full generation and forced answer completion. 

\begin{table*}[!htbp]
\centering
\small
\renewcommand{\arraystretch}{0.9}
\caption{Answer-trajectory Metrics computed across different datasets. Except for Answer Switches, lower values of all metrics indicate fewer redundant tokens generated after the final decision, and therefore greater efficiency. For $\Pr(t^\star=-1)$, we compute the fraction of examples with no answer switch on the raw traces and on the hold-for-$k$ denoised traces, respectively. For $T_{\mathrm{after}}$ each entry shows ``tokens (percentage of reasoning tokens)''.  We present detailed distributions of these metrics in Appendix~\ref{sec:boxplots}}.
\label{tab:trajectory-summary}
\begin{tabular}{l l c c c c}
      \toprule
      Dataset                       & Model                             &
      \shortstack{$\Pr(t^\star=-1)$                                                                                                            \\Raw\% / Denoised\%} &
      \shortstack{AnswerSwitches                                                                                                               \\Raw / Denoised} &
      \shortstack{TAF$(T)$} &
      \shortstack{$T_{\mathrm{after}}$                                                                                                         \\Raw / Denoised} \\
      \midrule

      \multirow{6}{*}{MCQ}
                                    & Qwen3-4B
                                    & 58.6 / 67.0
                                    & 3.59 / 0.61
                                    & 2.02
                                    & 555.2 (35.7\%) / 867.1 (43.1\%)                                                                          \\
                                    & Qwen3-8B
                                    & 50.5 / 59.2
                                    & 4.30 / 0.80
                                    & 2.29
                                    & 663.0 (35.9\%) / 1033.4 (43.4\%)                                                                         \\
                                    & Qwen3-14B
                                    & 55.4 / 65.3
                                    & 3.11 / 0.64
                                    & 1.64
                                    & 609.9 (35.4\%) / 783.5 (39.3\%)                                                                          \\
                                    & Qwen3-30B-A3B
                                    & 67.9 / 73.1
                                    & 2.17 / 0.44
                                    & 1.16
                                    & 641.5 (40.1\%) /  868.5(45.5\%)                                                                          \\
                                    & GPT-OSS-20B (High)                & 56.1 / 63.6 & 2.16 / 0.63  & 0.97 & 583.7 (47.0\%) / 712.6 (46.3\%)  \\
                                    & GPT-OSS-20B (Medium)              & 50.3 / 60.8 & 2.91 / 0.75  & 1.36 & 284.1 (35.7\%) / 361.2 (31.2\%)  \\
      \midrule

      \multirow{6}{*}{Numeric-answer}
                                    & Qwen3-4B
                                    & 19.4 / 23.5
                                    & 7.15 / 1.64
                                    & 3.40
                                    & 846.7 (42.6\%) / 1192.9 (51.8\%)                                                                         \\
                                    & Qwen3-8B
                                    & 20.8 / 24.8
                                    & 9.26 / 1.74
                                    & 4.68
                                    & 1280 (45.8\%) / 1904.4 (59.9\%)                                                                          \\
                                    & Qwen3-14B
                                    & 18.2 / 27.3
                                    & 8.77 / 1.80
                                    & 4.43
                                    & 1271.7 (50.5\%) / 1717.3 (62.4\%)                                                                        \\
                                    & Qwen3-30B-A3B
                                    & 31.2 / 37.5
                                    & 6.75 / 1.06
                                    & 3.75
                                    & 1203.8 (50.5\%) / 1830.4 (70.2\%)                                                                        \\
                                    & GPT-OSS-20B (High)                & 9.9 / 14.6  & 13.02 / 2.35 & 5.82 & 823.6 (39.5\%) / 1437.5 (51.9\%) \\
                                    & GPT-OSS-20B (Medium)              & 8.0 / 13.3  & 13.83 / 2.72 & 5.77 & 307.0 (25.4\%) / 466.8 (27.7\%)  \\
      \midrule

      \multirow{6}{*}{Search-query}
                                    & Qwen3-4B
                                    & 53.5 / 63.0
                                    & 0.85 / 0.41
                                    & 0.26
                                    & 148.1 (64.7\%) / 106.5 (44.7\%)                                                                          \\
                                    & Qwen3-8B
                                    & 60.1 / 74.5
                                    & 0.87 / 0.28
                                    & 0.38
                                    & 145.5 (58.6\%) / 121.6 (46.6\%)                                                                          \\
                                    & Qwen3-14B
                                    & 68.5 / 79.8
                                    & 0.61 / 0.22
                                    & 0.26
                                    & 114.4 (54.4\%) /  78.5 (36.3\%)                                                                          \\
                                    & Qwen3-30B-A3B
                                    & 60.1 / 80.2
                                    & 0.91 / 0.22
                                    & 0.40
                                    & 95.8 (45.2\%) / 82.2 (35.3\%)                                                                            \\
                                    & GPT-OSS-20B (High)                & 74.7 / 97.3 & 0.55 / 0.03  & 0.29 & 36.8 (19.4\%) / 63.4 (29.4\%)    \\
                                    & GPT-OSS-20B (Medium)              & 88.4 / 99.6 & 0.17 / 0.00  & 0.07 & 9.1 (15.5\%) / 12.4 (10.1\%)     \\
      \midrule

      \multirow{6}{*}{Tool-selection}
                                    & Qwen3-4B
                                    & 97.7 / 98.8
                                    & 0.21 / 0.02
                                    & 0.14
                                    & 647.0 (41.3\%) / 855.7 (56.6\%)                                                                          \\
                                    & Qwen3-8B
                                    & 97.4 / 98.6
                                    & 0.11 / 0.02
                                    & 0.06
                                    & 338 (44.9\%) / 465.3 (46.8\%)                                                                            \\
                                    & Qwen3-14B
                                    & 96.4 / 98.0
                                    & 0.16 / 0.03
                                    & 0.09
                                    & 331.6 (44.6\%) / 306.6 (38.5\%)                                                                          \\
                                    & Qwen3-30B-A3B
                                    & 99.0 / 100
                                    & 0.02 / 0.00
                                    & 0.01
                                    & 37.0 (23.6\%) / 37.0 (23.6\%)                                                                            \\
                                    & GPT-OSS-20B (High)                & 87.3 / 96.1 & 0.79 / 0.08  & 0.44 & 576.9 (53.1\%) / 677.8 (54.8\%)  \\
                                    & GPT-OSS-20B (Medium)              & 93.0 / 97.1 & 0.31 / 0.05  & 0.17 & 169.5 (49.6\%) / 147.2 (39.1\%)  \\
      \bottomrule
    \end{tabular}
\end{table*}

\section{Results}\label{sec:results}
We report results on how the model’s predicted answer evolves across reasoning steps using forced answer completion, and how these dynamics can be used to reduce the number of generated reasoning tokens. We analyze the frequency and distribution of answer switches and transient answer flips across tasks and models, and measure how many reasoning tokens are generated after the final answer switch. Building on these observations, we evaluate early stopping strategies that halt generation at intermediate reasoning steps and quantify the resulting trade-off between token usage and final answer accuracy.

\subsection{Properties of Answer Trajectories}
We discuss observed properties of answer trajectories obtained via forced answer completion, focusing on how predicted answers change across reasoning steps and how these changes are distributed across models and tasks.
\subsubsection{Transient Answer Flips}
We first examine the occurrence of transient answer flips, defined as short-lived deviations in the predicted answer that revert within a small number of reasoning steps. Across all evaluated models and task settings, transient answer flips are common, indicating frequent, brief changes in the predicted answer that do not persist. Table~\ref{tab:trajectory-summary} reports the average number of transient answer flips per example for each model across tasks. We observe that transient flips occur frequently, \textbf{$\approx$ 3 TAFs per reasoning trace}, across all models for multiple-choice and numeric answers. We find a lower transient answer flip count for tool-use and search which is consistent with the lower answer switches.
\subsubsection{Answer Trajectories with No Answer Switches}
Next, we analyze the fraction of examples in which the predicted answer does not change at any reasoning step, i.e., the answer trajectory satisfies \(A_0 = A_1 = \dots = A_n\) or $t^\star = -1$. In such cases, the predicted answer under forced completion remains identical to the non-reasoning baseline throughout the entire reasoning trace.

Figure~\ref{fig:answer-switch-distribution}
shows the distribution of the number of answer switches across examples. A substantial fraction of examples \textbf{($\approx$ 50\% examples in the MCQ dataset across all models)} exhibit zero answer switches, indicating that for these inputs the predicted answer is unchanged by the generated reasoning steps.

{\setlength{\textfloatsep}{6pt}
 \setlength{\floatsep}{6pt}
 \setlength{\intextsep}{6pt}

\begin{figure}[t]
    \centering
    \resizebox{.80\linewidth}{!}{%
        \input{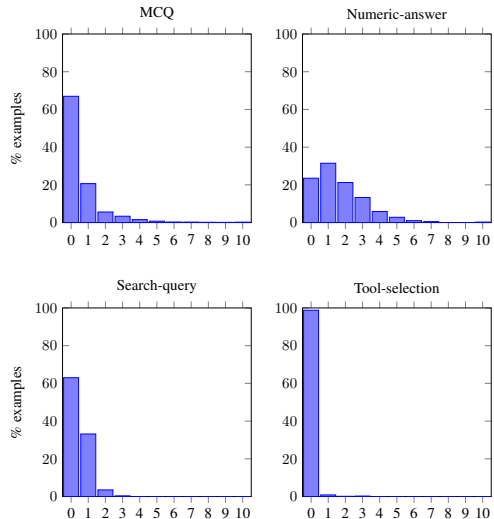}
    }
    \caption{Distribution of answer switch counts across denoised answer trajectories for different task types using Qwen3-4B.}
    \label{fig:answer-switch-distribution}
\end{figure}
}

\subsubsection{Limited Number of Answer Switches When Switches Occur}
We consider examples in which the predicted answer does change at least once. Among these examples, the total number of answer switches is typically small relative to the length of the reasoning trace.

Figure~\ref{fig:answer-switch-distribution}
presents the distribution of answer switches. The majority of such examples exhibit only a small number of switches, with long stretches of reasoning steps between successive changes.
\vspace{-0.2cm}

\subsubsection{Tool-Selection Tasks Exhibit Near-Zero Answer Switches}
For tool-selection datasets, from Table~\ref{tab:trajectory-summary} we observe that the predicted answer is highly stable across reasoning steps as seen from $\approx 0$ Answer Switches for the Tool Selection Dataset across all models. In the vast majority of examples, the model selects the same tool from the initial forced completion onward, resulting in almost no answer switches throughout the reasoning trace. Consequently, early stopping at intermediate steps does not alter the selected tool in practice.

Because the predicted tool remains unchanged, early stopping induces no measurable degradation in task performance. For this reason, we do not report accuracy–token trade-off curves for tool-selection tasks, as the accuracy is effectively invariant to the stopping point under our evaluation protocol.

Despite this stability, we include hidden states from tool-selection examples when training the generic probe. Although these examples provide limited signal for detecting answer switches, they expose the probe to reasoning representations from a distinct task regime, improving robustness and cross-task generalization. In particular, incorporating tool-selection data helps the generic probe learn representations associated with early decision commitment.

\subsection{Tokens After Final Answer Switch}

We now analyze how many reasoning tokens are generated after the model’s final answer switch, as defined in Section~\ref{sec:final_answer_switch}. For each example, we measure the number of tokens generated after the last reasoning step at which the predicted answer changes.

Table~\ref{tab:trajectory-summary}
shows the average number of tokens generated after the final answer switch per example. We observe that a large fraction of the total reasoning tokens 
\textbf{($\approx $50\% of reasoning tokens across all dataset and models)} are generated after the predicted answer has already reached its final value.

These tokens represent reasoning that does not alter the model’s predicted answer under forced completion. As a result, the measured quantity provides an upper bound on the number of tokens that could be avoided by an ideal early stopping strategy that halts generation immediately after the final answer switch.

While practical early stopping methods may not perfectly identify this point, the observed distributions indicate that a substantial portion of generated reasoning tokens occur after the final answer switch, leaving significant room for reducing token usage without changing the predicted answer.

\subsection{Evaluation on Harder Reasoning Benchmarks}
We extend our analysis to substantially harder reasoning benchmarks to evaluate whether the observed answer trajectory dynamics persist in more challenging settings. Specifically, we consider Humanity’s Last Exam~\cite{phan2025lastexam}, GPQA-Diamond~\cite{rein2023gpqagraduatelevelgoogleproofqa}, and AIME 2026 problems~\cite{balunovic_srimatharena_2025}, which are designed to require deeper multi-step reasoning and exhibit lower overall accuracy compared to the datasets considered earlier.

Across all three benchmarks, we observe qualitatively consistent behavior with our main results. In particular, the predicted answer often stabilizes before the completion of the full reasoning trace, even in these more demanding settings. Although these datasets exhibit higher answer-switch counts and require longer reasoning chains, the final answer switch typically occurs well before the end of generation.

We measure the number of reasoning tokens generated after the final answer switch as defined in Section 3.5. Despite the increased task difficulty, a substantial fraction of reasoning tokens are generated after the predicted answer has already stabilized. On average, we find that approximately 32\%--67\% of the total reasoning tokens are produced after the final answer switch across these datasets.

These results indicate that the temporal gap between decision formation and subsequent reasoning is not limited to moderate-difficulty or proxy tasks, but persists even in substantially harder reasoning regimes. This further supports the generality of the early decision-making phenomenon observed in our study. We list the metrics calculated in Appendix~\ref{app:aux-datasets}

\subsection{Effect of Denoising}
We evaluate the effect of answer trajectory denoising on the structure of predicted answer sequences. The denoising procedure is applied to suppress transient answer flips while preserving sustained answer changes.

\begin{figure}
    \centering
    \resizebox{.8\linewidth}{!}{%
        \input{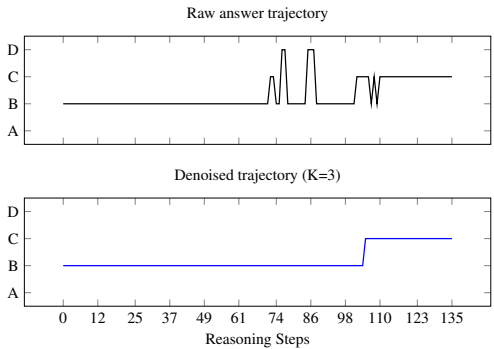}
    }
    \caption{Illustration of Answer Switch Denoising for MCQ task. After denoising, the decision path becomes clearer across the entire reasoning process.}
    \label{fig:denoising_example}
\end{figure}

Figure~\ref{fig:denoising_example} compares the original and denoised answer trajectory of an example: Short-lived answer changes present in the raw trajectory are removed after denoising, while longer-lasting answer changes are retained.

We also show the effect of denoising at the dataset level in by the distribution plots in Appendix \ref{sec:boxplots}. Compared to the original distribution, the denoised distribution exhibits fewer examples with a large number of answer switches, and the mass is concentrated on smaller switch counts. For example, as seen in Table~\ref{tab:trajectory-summary}, the average answer switches decrease when the trajectory is denoised, \textbf{$\mathbf{\approx 8 \to \approx 2}$ for the Numeric-answer dataset across all models}.

These results show that denoising removes transient answer flips and yields answer switch distributions that are less spread out, while preserving the overall structure of sustained answer changes.
\vspace{-0.2cm}

\section{Early Stopping and Performance Trade-Offs}
\label{sec:early_stopping}

We evaluate two early-stopping strategies for determining when to terminate a model’s reasoning process, i.e., when intermediate predictions are redundant to the final decision.


\subsection{Early Stopping Gates}
We introduce early-stopping gates that operate over partial reasoning prefixes and produce a scalar confidence score indicating whether generation can be terminated at a given step. These gates do not alter the model’s generation process; rather, they observe intermediate states and define stopping rules based on predefined criteria.

Formally, at each reasoning step $i \in \{0,\ldots,n\}$, a gate observes information available up to that step and outputs a scalar score:
\begin{equation}\small
m_i = \mathcal{M}(C, R_1,\ldots,R_i),
\qquad m_i \in [0,1],    
\end{equation}

where $\mathcal{M}$ denotes the gate function. The score $m_i$ is interpreted as the gate’s confidence that further reasoning is unnecessary.

Given a threshold $\tau \in [0,1]$, the corresponding stopping rule halts generation at the earliest step
\begin{equation}\small
s = \min \{\, i \mid m_i \ge \tau \,\}.
\end{equation}
Varying $\tau$ allows explicit control over the tradeoff between final answer accuracy and token savings.

We consider the two gating criteria, random gating and Probe-Based gate. The \textbf{random gate} outputs a confidence score drawn independently at each reasoning step. Specifically, at step $i$ the gate produces $m_i \sim \mathrm{Uniform}(0,1)$.

The \textbf{probe-based gate} uses a learned linear transformation over a layer's intermediate model representations to estimate whether further reasoning is necessary. Let $h^l_i \in \mathbb{R}^d$ denote the final token hidden state extracted from the model's $l$th layer at reasoning step $i$. The gate computes
\begin{equation}\small
m_i = \sigma(w^\top h^l_i + b),
\end{equation}

where $w \in \mathbb{R}^d$, $b \in \mathbb{R}$ are learned parameters and $\sigma(\cdot)$ denotes the sigmoid function, ensuring $m_i \in (0,1)$. The gate is trained to predict whether the predicted answer at step $i$ matches the eventual final answer produced at step $n$. The supervision signal is derived post hoc from full reasoning traces and does not influence the generation process.

\subsection{Experimental Setup}

\paragraph{Reasoning trace collection.}
For each dataset \(D\), we first collect full reasoning traces by running the model to completion. For each example, this yields a reasoning trace \(R = (R_1,\dots,R_n)\). We then use forced answer completion to get an answer trajectory \(T = (A_0,\dots,A_n)\), and the index of the final answer switch \(t^\star\) as defined in Section~\ref{sec:final_answer_switch}. Collected traces are split at the example level into training, validation, and test sets. All probe parameters are learned using only the training split, with hyperparameters selected on the validation split. The test split is held out for final evaluation only.


\begin{figure*}[!htbp]
    \centering
    \resizebox{.93\linewidth}{!}{%
        \input{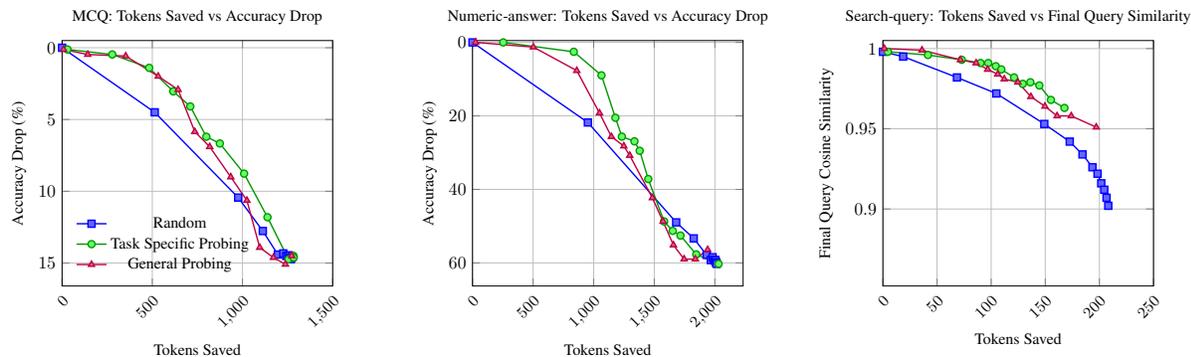}
    }
    \caption{Early stopping performance for random, task-specific and generic probe-based gates on Qwen3-4B reasoning.}
    \label{fig:early_stopping_perf}
\end{figure*}

\paragraph{Task-specific and generic probe-based gate.}
We train probe-based gates under two settings. In the task-specific setting, we train a separate probe for each dataset using only that dataset’s training split, allowing the probe to specialize to dataset-specific reasoning dynamics. In addition, we train a generic probe on the union of training splits from all datasets. The motivation is to test whether there exists a shared, task-agnostic structure in the model’s intermediate representations that reliably signals when further reasoning is unnecessary. The generic probe is evaluated on each dataset’s test split without any dataset-specific adaptation. Comparing task-specific and generic probes, therefore, we examine whether early stopping can be governed by a universal signal, rather than task or dataset-specific heuristics. Further training details are given in Appendix~\ref{sec:training_details}.

\subsection{Early Stopping Setup}
We plot the number of reasoning tokens saved and the resulting drop in final answer accuracy for MCQ and Numeric test sets in the left two of  Figure~\ref{fig:early_stopping_perf}, and the cosine similarity of the intermediate query and the final query for the Search test set in the right one.

First, probe-based gates in both task-specific and generic settings can effectively \textbf{save 500 tokens with less than 2\% accuracy drop} on Qwen3-4B. At higher savings of 1,000 tokens, generic probing on the numeric dataset still limits the accuracy drop to roughly 4\%. Although numeric tasks are inherently more sensitive to truncated reasoning, probing continues to provide a markedly better trade-off, suggesting that answer determination often precedes the completion of full arithmetic derivations.

For the search task, we measure degradation using final query cosine similarity rather than accuracy. Generic probing preserves a cosine similarity of $\approx 0.98$ while saving up to 100 tokens, and remains above \textbf{0.95 even when saving 200 tokens}. In contrast, random stopping drops below 0.93 at similar token savings. This demonstrates that probing-based early stopping preserves the semantic content of the final query even when a substantial fraction of reasoning tokens is removed.

Overall, these quantitative results align with our trajectory analysis, showing that, on Qwen3-4B predicted answers change in only 32\% of queries and that models generate an average of 760 additional tokens after the final answer is determined, averaged across all tasks. Probing-based early stopping capitalizes on this redundancy, \textbf{saving up to 500–900 tokens per query with only 1–5\% accuracy degradation} in many regimes. While we do not observe a consistent trend with respect to model size, similar patterns emerge across different scales. Collectively, this confirms that a significant portion of chain-of-thought generation is unnecessary for correctness and can be safely eliminated using simple, inference-time heuristics. Exhaustive results are presented in Section~\ref{app:full_probing_results} in the Appendix.
\vspace{-0.375cm}
\paragraph{Individual vs Generic Probe-based gate} Across tasks, generic probe closely matches token–performance trade-off achieved by individually trained task-specific gates, with no consistent loss in token savings or final answer quality.

Despite being trained without any dataset-specific supervision, the generic probe gate yields early stopping frontiers that largely overlap with those of the individual probes in Figure~\ref{fig:early_stopping_perf}. This indicates that the intermediate representations used by the probe to decide when further reasoning is unnecessary capture signals that are largely shared across tasks. The comparable performance of the generic and individual gates therefore suggests the presence of a larger universal structure governing answer switches and early stopping, rather than task-specific stopping behavior.

\section{Discussion and Limitations}
Our results show that, across models and tasks, our predicted answer often stabilizes early in the reasoning process. Many examples show no answer switches at all, and when the switches do occur, their number is small. Importantly, a large portion of reasoning tokens are generated after the final answer switch, meaning they do not change the predicted answer under forced completion. This explains why early stopping can save a substantial number of tokens with only a small drop in performance. Probe-based stopping is especially effective, and the strong performance of the generic probe suggests that models encode shared signals that indicate when further reasoning is unlikely to change the answer.

Despite its promising results, our study is limited to the models, tasks, and reasoning lengths evaluated in this work. We do not study substantially larger models or much longer reasoning traces, where answer dynamics may differ. In addition, our experiments focus on language based reasoning tasks, and we leave extensions to broader multimodal models for future work.

\section{Impact Statement}
This work presents an inference-time approach for reducing unnecessary reasoning generation in large language models. By stopping generation once the answer has stabilized, the method lowers token usage, inference cost, and latency without modifying the base model. These efficiency gains can help reduce the computational and environmental footprint of deploying reasoning models at scale.

\bibliographystyle{icml2026}
\bibliography{example_paper}

\newpage
\appendix

\section{Datasets}

We use the sample sizes described in Table~\ref{tab:sample-size} to conduct our analyses and early stopping experiments for all the datasets. We shuffle and randomly sample from the dataset to obtain samples.

\begin{table}[H]
    \centering
    \begin{tabular}{cc}
    \hline

        Dataset & $S$ \\
\hline

         MCQ & 1000 \\
         Numeric-answer & 500 \\
         Search-query & 1000 \\
         Tool-selection & 1000 \\
\hline

    \end{tabular}
    \caption{Sample Sizes of reasoning traces}
    \label{tab:sample-size}
\end{table}

\section{Answer Prefix}\label{app:answer_prefix}

We choose an answer prefix $P$ that elicits a probability distribution over the task answer space from the model’s vocabulary-level output distribution. The choice of $P$ depends on the task format. Below, we describe the prefixes used for each class of task considered in this work.

\subsection{MCQ Answer Prefix}

For multiple-choice questions (MCQs), we instruct the model to respond in a fixed JSON format:
\texttt{\{"answer": "A"\}}, where \texttt{A} can be replaced by any valid answer label.
During forced decoding, we set the answer prefix to
$P = \texttt{\{"answer": "}$,
which induces a distribution over the possible MCQ answer labels at the next token position.
We intervene at each reasoning step and record the unnormalized logits corresponding to the candidate answer labels.

\begin{table}[h]
\centering
\begin{tabular}{ll}
\hline
\textbf{Component} & \textbf{Value} \\
\hline
Expected output format & \texttt{\{"answer": "A"\}} \\
Answer prefix $P$ & \texttt{\{"answer": "} \\
Predicted token & A, B, C or D\\
\hline
\end{tabular}
\caption{Answer prefix used for MCQ tasks.}
\end{table}

\subsection{Numeric Answer Prefix}

For tasks whose answers are numeric, we instruct the model to output its answer inside
\texttt{\textbackslash boxed\{\}} tags, e.g., \texttt{\textbackslash boxed\{42\}}.
During forced decoding, we set
$P = \texttt{\textbackslash boxed\{}$
to elicit a distribution over possible numeric answers.
We restrict our analysis to questions whose correct answer is represented by a single token in the model vocabulary.

\begin{table}[h]
\centering
\begin{tabular}{ll}
\hline
\textbf{Component} & \textbf{Value} \\
\hline
Expected output format & \texttt{\textbackslash boxed\{42\}} \\
Answer prefix $P$ & \texttt{\textbackslash boxed\{} \\
Predicted token & Single-token numeric answer \\
\hline
\end{tabular}
\caption{Answer prefix used for numeric-answer tasks.}
\end{table}

\subsection{Tool Selection Prefix}

For user requests where the model must select exactly one tool from a provided list, we track the evolution of the selected tool across reasoning steps.
To simplify the setup, we consider tools without arguments.
We use the model’s native tool-calling format and choose a prefix such that the next generated token corresponds to the tool name.

For models using Hermes-style function calling (e.g., Qwen3), the format is:

\begin{table}[h]
\centering
\begin{tabular}{p{0.95\linewidth}}
\hline
\textbf{Prefix Format} \\
\hline
\texttt{C<think>R$_1$R$_2$\ldots R$_k$</think>} \\
\texttt{<tool\_call>\{"name": "} \\

\hline
\end{tabular}
\caption{Answer prefix used for tool selection tasks. The highlighted token corresponds to the predicted tool name.}
\end{table}

\subsection{Tool Argument Prefix}

For tasks where the model is forced to call a specific tool that takes a single argument (e.g., a web search tool), we track the evolution of the generated argument across reasoning steps.
Generation proceeds using the same sampling settings and is terminated at the closing tool call tag (\texttt{</tool\_call>}).

The prefix used to elicit the argument token is shown below.

\begin{table}[h]
\centering
\begin{tabular}{p{0.95\linewidth}}
\hline
\textbf{Prefix Format} \\
\hline
\texttt{C<think>R$_1$R$_2$\ldots R$_k$</think>} \\
\texttt{<tool\_call>\{"name": "web\_search", "arguments": \{"query": "} \\
\hline
\end{tabular}
\caption{Answer prefix used for tool argument generation. The highlighted token corresponds to the generated argument.}
\end{table}

\section{Full Early Stopping Results}\label{app:full_probing_results}

We evaluate early stopping methods on Qwen/Qwen3-4B and Qwen/Qwen3-8B.

\subsection{Qwen/Qwen3-4B}
Tables~\ref{tab:mcq-4b}-\ref{tab:s-q-4b-general} show the results of early stopping experiments for Qwen/Qwen3-4B

\subsubsection{Random probing}
\begin{table}[H]
\centering
\caption{MCQ}
\label{tab:mcq-4b}
\begin{tabular}{lrr}
\hline
Accuracy drop & Tokens saved & Token \%\\\hline
14.620 & 1273 & 99.2\\
14.596 & 1271 & 99.0\\
14.725 & 1268 & 98.8\\
14.678 & 1264 & 98.5\\
14.480 & 1259 & 98.1\\
14.468 & 1253 & 97.6\\
14.515 & 1243 & 96.8\\
14.327 & 1226 & 95.5\\
14.409 & 1196 & 93.2\\
12.772 & 1113 & 86.7\\
10.444 & 976 & 76.1\\
4.503 & 513 & 40.0\\
0.000 & 0 & 0.0\\
\hline\end{tabular}
\end{table}

\begin{table}[H]
\centering
\caption{Numeric-answer}
\label{n-a-4b}
\begin{tabular}{lrr}
\hline
Accuracy drop & Tokens saved & Token \%\\\hline
60.256 & 2014 & 99.2\\
59.872 & 2012 & 99.1\\
59.872 & 2008 & 98.9\\
59.231 & 2005 & 98.7\\
59.359 & 2000 & 98.5\\
59.231 & 1993 & 98.2\\
58.462 & 1981 & 97.6\\
59.231 & 1966 & 96.8\\
57.821 & 1932 & 95.2\\
53.333 & 1825 & 89.9\\
48.974 & 1681 & 82.8\\
21.795 & 952 & 46.9\\
0.000 & 0 & 0.0\\
\hline\end{tabular}
\end{table}

\begin{table}[H]
\centering
\caption{Search-query}
\label{s-q-4b}
\begin{tabular}{lrr}
\hline
Cosine similarity & Tokens saved & Token \%\\\hline
0.902 & 208 & 91.0\\
0.907 & 207 & 90.3\\
0.912 & 205 & 89.4\\
0.916 & 202 & 88.2\\
0.922 & 198 & 86.7\\
0.926 & 194 & 84.7\\
0.934 & 184 & 80.6\\
0.942 & 173 & 75.5\\
0.953 & 149 & 65.2\\
0.972 & 105 & 45.8\\
0.982 & 68 & 29.9\\
0.995 & 19 & 8.2\\
0.998 & 0 & 0.0\\
\hline\end{tabular}
\end{table}

\subsubsection{Task-specific probing}
\begin{table}[H]
\centering
\caption{MCQ}
\begin{tabular}{lrr}
\hline
Accuracy drop & Tokens saved & Token \%\\\hline
14.620 & 1283 & 100.0\\
14.620 & 1283 & 100.0\\
14.503 & 1283 & 99.9\\
14.737 & 1255 & 97.8\\
11.813 & 1139 & 88.7\\
8.772 & 1009 & 78.6\\
6.667 & 874 & 68.1\\
6.199 & 800 & 62.3\\
4.094 & 711 & 55.4\\
3.041 & 616 & 48.0\\
1.404 & 483 & 37.6\\
0.468 & 278 & 21.7\\
0.117 & 30 & 2.3\\
\hline\end{tabular}
\end{table}

\begin{table}[H]
\centering
\caption{Numeric-answer}
\begin{tabular}{lrr}
\hline
Accuracy drop & Tokens saved & Token \%\\\hline
60.256 & 2030 & 100.0\\
57.692 & 1846 & 90.9\\
52.564 & 1716 & 84.5\\
51.282 & 1652 & 81.4\\
48.718 & 1581 & 77.9\\
37.179 & 1450 & 71.4\\
29.487 & 1380 & 68.0\\
26.923 & 1336 & 65.8\\
25.641 & 1233 & 60.7\\
20.513 & 1179 & 58.1\\
8.974 & 1064 & 52.4\\
2.564 & 836 & 41.2\\
0.000 & 254 & 12.5\\
\hline\end{tabular}
\end{table}

\begin{table}[H]
\centering
\caption{Search-query}
\begin{tabular}{lrr}
\hline
Cosine similarity & Tokens saved & Token \%\\\hline
0.963 & 168 & 73.3\\
0.968 & 155 & 67.9\\
0.977 & 144 & 63.2\\
0.979 & 136 & 59.6\\
0.978 & 129 & 56.5\\
0.982 & 121 & 53.0\\
0.987 & 109 & 47.8\\
0.989 & 104 & 45.6\\
0.991 & 97 & 42.6\\
0.991 & 90 & 39.4\\
0.993 & 73 & 31.8\\
0.996 & 42 & 18.2\\
0.998 & 5 & 2.0\\
\hline\end{tabular}
\end{table}

\subsubsection{General probing}
\begin{table}[H]
\centering
\caption{MCQ}
\begin{tabular}{lrr}
\hline
Accuracy drop & Tokens saved & Token \%\\\hline
14.503 & 1273 & 99.2\\
15.088 & 1238 & 96.4\\
14.620 & 1171 & 91.3\\
13.918 & 1096 & 85.4\\
10.643 & 1025 & 79.8\\
9.006 & 935 & 72.9\\
6.901 & 818 & 63.8\\
5.848 & 736 & 57.3\\
2.924 & 643 & 50.1\\
1.988 & 533 & 41.5\\
0.585 & 353 & 27.5\\
0.468 & 142 & 11.1\\
0.117 & 8 & 0.6\\
\hline\end{tabular}
\end{table}

\begin{table}[H]
\centering
\caption{Numeric-answer}
\begin{tabular}{lrr}
\hline
Accuracy drop & Tokens saved & Token \%\\\hline
56.410 & 1940 & 95.6\\
58.974 & 1838 & 90.6\\
58.974 & 1746 & 86.0\\
55.128 & 1658 & 81.7\\
48.718 & 1572 & 77.4\\
42.308 & 1484 & 73.1\\
30.769 & 1298 & 63.9\\
28.205 & 1249 & 61.5\\
25.641 & 1146 & 56.4\\
19.231 & 1048 & 51.6\\
7.692 & 861 & 42.4\\
1.282 & 502 & 24.7\\
0.000 & 25 & 1.2\\
\hline\end{tabular}
\end{table}

\begin{table}[H]
\centering
\caption{Search-query}
\begin{tabular}{lrr}
\hline
Cosine similarity & Tokens saved & Token \%\\\hline
0.951 & 197 & 86.2\\
0.958 & 174 & 76.1\\
0.958 & 161 & 70.4\\
0.964 & 150 & 65.4\\
0.970 & 137 & 59.8\\
0.979 & 125 & 54.5\\
0.981 & 112 & 49.0\\
0.984 & 106 & 46.4\\
0.987 & 97 & 42.3\\
0.991 & 86 & 37.7\\
0.993 & 72 & 31.3\\
0.999 & 36 & 15.8\\
1.000 & 1 & 0.6\\
\hline\end{tabular}
\label{tab:s-q-4b-general}
\end{table}

\begin{figure*}[!htbp]
    \centering
    \resizebox{.95\linewidth}{!}{%
        \input{figures/Qwen3-8B/early_stopping_performance}
    }
    \caption{Early stopping performance for random, task-specific and generic probe-based gates on Qwen3-8B reasoning.}
    \label{fig:early_stopping_perf_8B}
\end{figure*}

\subsection{Qwen/Qwen3-8B}
Tables~\ref{tab:mcq-8b-random}-\ref{tab:s-q-8b-combined} show the results of early stopping experiments for Qwen/Qwen3-8B. We plot these results in Figure~\ref{fig:early_stopping_perf_8B}.

\subsubsection{Random probing}
\begin{table}[H]
\centering
\caption{MCQ}\begin{tabular}{lrr}
\hline
Accuracy drop & Tokens saved & Token \%\\\hline
24.390 & 1764 & 99.5\\
24.429 & 1762 & 99.4\\
24.527 & 1759 & 99.2\\
24.527 & 1755 & 99.0\\
24.473 & 1751 & 98.8\\
24.571 & 1745 & 98.4\\
24.732 & 1735 & 97.9\\
24.707 & 1721 & 97.0\\
24.141 & 1690 & 95.3\\
22.459 & 1608 & 90.7\\
19.107 & 1468 & 82.8\\
9.112 & 854 & 48.2\\
0.488 & 0 & 0.0\\
\hline\end{tabular}
\label{tab:mcq-8b-random}
\end{table}

\begin{table}[H]
\centering
\caption{Numeric-answer}
\begin{tabular}{lrr}
\hline
Accuracy drop & Tokens saved & Token \%\\\hline
59.722 & 2920 & 99.3\\
59.306 & 2918 & 99.3\\
59.028 & 2914 & 99.1\\
59.306 & 2910 & 99.0\\
59.167 & 2905 & 98.8\\
58.611 & 2898 & 98.6\\
58.333 & 2886 & 98.2\\
59.444 & 2872 & 97.7\\
58.750 & 2839 & 96.6\\
55.000 & 2734 & 93.0\\
50.556 & 2559 & 87.1\\
24.792 & 1588 & 54.0\\
0.000 & 0 & 0.0\\
\hline\end{tabular}
\end{table}

\begin{table}[H]
\centering
\caption{Search-query}
\begin{tabular}{lrr}
\hline
Cosine similarity & Tokens saved & Token \%\\\hline
0.927 & 257 & 92.5\\
0.929 & 256 & 92.0\\
0.932 & 254 & 91.2\\
0.935 & 251 & 90.2\\
0.939 & 247 & 89.0\\
0.943 & 242 & 87.2\\
0.945 & 233 & 83.7\\
0.949 & 220 & 79.2\\
0.959 & 196 & 70.4\\
0.971 & 139 & 50.0\\
0.979 & 94 & 33.9\\
0.992 & 25 & 8.9\\
0.996 & 0 & 0.0\\
\hline\end{tabular}
\end{table}

\subsubsection{Task-specific probing}
\begin{table}[H]
\centering
\caption{MCQ}
\begin{tabular}{lrr}
\hline
Accuracy drop & Tokens saved & Token \%\\\hline
23.415 & 1763 & 99.4\\
25.366 & 1673 & 94.4\\
24.878 & 1617 & 91.2\\
25.854 & 1552 & 87.5\\
20.976 & 1457 & 82.2\\
14.634 & 1350 & 76.1\\
10.244 & 1211 & 68.3\\
6.829 & 1100 & 62.1\\
3.902 & 993 & 56.0\\
0.976 & 843 & 47.5\\
1.951 & 545 & 30.7\\
0.488 & 279 & 15.7\\
0.000 & 34 & 1.9\\
\hline\end{tabular}
\end{table}

\begin{table}[H]
\centering
\caption{Numeric-answer}
\begin{tabular}{lrr}
\hline
Accuracy drop & Tokens saved & Token \%\\\hline
59.722 & 2939 & 100.0\\
56.250 & 2839 & 96.6\\
60.417 & 2595 & 88.3\\
55.556 & 2528 & 86.0\\
54.861 & 2458 & 83.6\\
50.000 & 2380 & 81.0\\
45.139 & 2282 & 77.6\\
41.667 & 2228 & 75.8\\
36.111 & 2185 & 74.3\\
31.944 & 2073 & 70.5\\
23.611 & 1962 & 66.7\\
15.278 & 1777 & 60.5\\
2.778 & 1311 & 44.6\\
\hline\end{tabular}
\end{table}

\begin{table}[H]
\centering
\caption{Search-query}
\begin{tabular}{lrr}
\hline
Cosine similarity & Tokens saved & Token \%\\\hline
0.946 & 248 & 89.2\\
0.956 & 210 & 75.5\\
0.957 & 197 & 70.7\\
0.960 & 185 & 66.5\\
0.967 & 172 & 61.8\\
0.966 & 164 & 58.9\\
0.972 & 150 & 53.8\\
0.972 & 142 & 51.1\\
0.975 & 136 & 48.9\\
0.979 & 126 & 45.4\\
0.979 & 113 & 40.6\\
0.985 & 96 & 34.6\\
0.994 & 50 & 18.0\\
\hline\end{tabular}
\end{table}

\subsubsection{General probing}
\begin{table}[H]
\centering
\caption{MCQ}
\begin{tabular}{lrr}
\hline
Accuracy drop & Tokens saved & Token \%\\\hline
23.902 & 1737 & 98.0\\
24.878 & 1695 & 95.6\\
23.902 & 1635 & 92.2\\
23.902 & 1578 & 89.0\\
22.439 & 1523 & 8app:aux-datasets5.9\\
17.073 & 1449 & 81.7\\
11.707 & 1313 & 74.1\\
8.293 & 1244 & 70.2\\
7.317 & 1163 & 65.6\\
6.829 & 1023 & 57.7\\
3.902 & 815 & 46.0\\
1.463 & 425 & 24.0\\
0.000 & 16 & 0.9\\
\hline\end{tabular}
\end{table}

\begin{table}[H]
\centering
\caption{Numeric-answer}
\begin{tabular}{lrr}
\hline
Accuracy drop & Tokens saved & Token \%\\\hline
61.039 & 2593 & 100.0\\
61.039 & 2593 & 100.0\\
60.390 & 2591 & 99.9\\
57.143 & 2432 & 93.8\\
46.753 & 2169 & 83.7\\
38.961 & 2068 & 79.8\\
33.117 & 1964 & 75.7\\
26.623 & 1813 & 69.9\\
22.727 & 1722 & 66.4\\
16.883 & 1660 & 64.0\\
11.039 & 1462 & 56.4\\
5.195 & 1072 & 41.4\\
0.000 & 238 & 9.2\\
\hline\end{tabular}
\end{table}

\begin{table}[H]
\centering
\caption{Search-query}
\begin{tabular}{lrr}
\hline
Cosine similarity & Tokens saved & Token \%\\\hline
0.956 & 237 & 85.4\\
0.958 & 217 & 77.9\\
0.961 & 200 & 72.1\\
0.959 & 186 & 66.8\\
0.965 & 173 & 62.1\\
0.967 & 166 & 59.6\\
0.963 & 154 & 55.5\\
0.967 & 147 & 52.8\\
0.975 & 134 & 48.2\\
0.975 & 121 & 43.7\\
0.982 & 88 & 31.8\\
0.990 & 39 & 14.0\\
0.997 & 2 & 0.8\\
\hline\end{tabular}
\label{tab:s-q-8b-combined}
\end{table}

\section{Answer Trajectory Metrics}
We plot the distributions of Answer Switches, Transient Answer Flips, and the Tokens after final switch as defined in Section~\ref{sec:answer-trajectory-metrics}

\label{sec:boxplots}

\subsection{Qwen3-4B}
\begin{figure}[H]
    \centering
    \resizebox{\linewidth}{!}{%
        \input{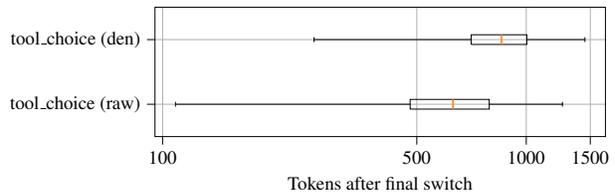}
    }
    \caption{Distribution of the mean of tokens after final answer switch for raw and denoised outputs in Qwen3-4B, evaluated on the Tool Selection.}
    \label{fig:tool_choice_tokens_afters_qwen4B}
\end{figure}

Figures \ref{fig:mmlu_ans_switch_qwen4B}-\ref{fig:tool_choice_tokens_afters_qwen4B} show distribution of metrics on Qwen3-8B.

\subsubsection{MCQ}

\begin{figure}[H]
    \centering
    \resizebox{\linewidth}{!}{%
        \input{figures/Qwen3-4B/mmlu_pro_bootstrap_metrics_answer_switches}
    }
    \caption{Distribution of the mean of answer switches for raw and denoised outputs in Qwen3-4B, evaluated on the MCQ Questions.}
    \label{fig:mmlu_ans_switch_qwen4B}
\end{figure}

\begin{figure}[H]
    \centering
    \resizebox{\linewidth}{!}{%
        \input{figures/Qwen3-4B/mmlu_pro_bootstrap_metrics_transient}
    }
    \caption{Distribution of the mean of transient flips for raw and denoised outputs in Qwen3-4B, evaluated on the MCQ Questions.}
    \label{fig:mmlu_taf_qwen4B}
\end{figure}

\begin{figure}[H]
    \centering
    \resizebox{\linewidth}{!}{%
        \input{figures/Qwen3-4B/mmlu_pro_bootstrap_metrics_tokens_after}
    }
    \caption{Distribution of the mean of tokens after final answer switch for raw and denoised outputs in Qwen3-4B, evaluated on the MCQ Questions.}
    \label{fig:mmlu_tokens_after_qwen4B}
\end{figure}

\subsubsection{Numeric}

\begin{figure}[H]
    \centering
    \resizebox{\linewidth}{!}{%
        \input{figures/Qwen3-4B/numeric_bootstrap_metrics_answer_switches}
    }
    \caption{Distribution of the mean of answer switches for raw and denoised outputs in Qwen3-4B, evaluated on the Numeric Answers.}
    \label{fig:numeric_ans_switch_qwen4B}
\end{figure}

\begin{figure}[H]
    \centering
    \resizebox{\linewidth}{!}{%
        \input{figures/Qwen3-4B/numeric_bootstrap_metrics_transient}
    }
    \caption{Distribution of the mean of transient flips for raw and denoised outputs in Qwen3-4B, evaluated on the Numeric Answers.}
    \label{fig:numeric_taf_qwen4B}
\end{figure}

\begin{figure}[H]
    \centering
    \resizebox{\linewidth}{!}{%
        \input{figures/Qwen3-4B/numeric_bootstrap_metrics_tokens_after}
    }
    \caption{Distribution of the mean of tokens after final answer switch for raw and denoised outputs in Qwen3-4B, evaluated on the Numeric Answers.}
    \label{fig:numeric_token_after_qwen4B}
\end{figure}

\subsubsection{Search}

\begin{figure}[H]
    \centering
    \resizebox{\linewidth}{!}{%
        \input{figures/Qwen3-4B/search_bootstrap_metrics_answer_switches}
    }
    \caption{Distribution of the mean of answer switches for raw and denoised outputs in Qwen3-4B, evaluated on the Search Query.}
    \label{fig:search_answer_switch_qwen4B}
\end{figure}

\begin{figure}[H]
    \centering
    \resizebox{\linewidth}{!}{%
        \input{figures/Qwen3-4B/search_bootstrap_metrics_transient}
    }
    \caption{Distribution of the mean of transient flips for raw and denoised outputs in Qwen3-4B, evaluated on the Search Query.}
    \label{fig:search_taf_qwen4B}
\end{figure}

\begin{figure}[H]
    \centering
    \resizebox{\linewidth}{!}{%
        \input{figures/Qwen3-4B/search_bootstrap_metrics_tokens_after}
    }
    \caption{Distribution of the mean of tokens after final answer switch for raw and denoised outputs in Qwen3-4B, evaluated on the Search Query.}
    \label{fig:search_taf_qwen4B}
\end{figure}

\subsubsection{Tool Selection}
\begin{figure}[H]
    \centering
    \resizebox{\linewidth}{!}{%
        \input{figures/Qwen3-4B/tool_choice_bootstrap_metrics_answer_switches}
    }
    \caption{Distribution of the mean of answer switches for raw and denoised outputs in Qwen3-4B, evaluated on the Tool Selection.}
    \label{fig:tool_choice_ans_switch_qwen4B}
\end{figure}

\begin{figure}[H]
    \centering
    \resizebox{\linewidth}{!}{%
        \input{figures/Qwen3-4B/tool_choice_bootstrap_metrics_transient}
    }
    \caption{Distribution of the mean of transient flips for raw and denoised outputs in Qwen3-4B, evaluated on the Tool Selection.}
    \label{fig:tool_choice_taf_qwen4B}
\end{figure}

\subsection{Qwen3-8B}
\begin{figure}[H]
    \centering
    \resizebox{\linewidth}{!}{%
        \input{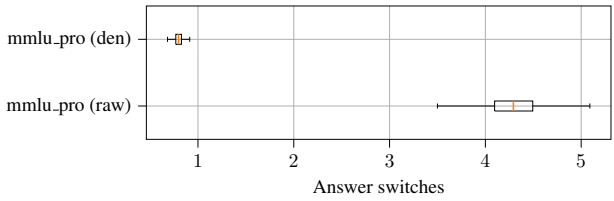}
    }
    \caption{Distribution of the mean of answer switches for raw and denoised outputs in Qwen3-8B, evaluated on the MCQ Questions.}
    \label{fig:mmlu_ans_switch_qwen8b}
\end{figure}

\begin{figure}[H]
    \centering
    \resizebox{\linewidth}{!}{%
        \input{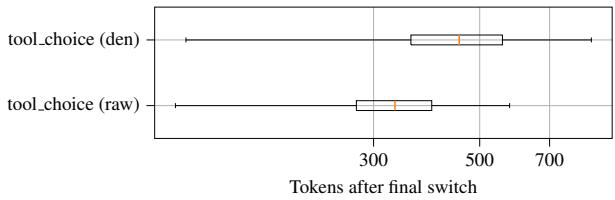}
    }
    \caption{Distribution of the mean of tokens after final answer switch for raw and denoised outputs in Qwen3-8B, evaluated on the Tool Selection.}
    \label{fig:tool_choice_tokens_afters_qwen8B}
\end{figure}

Figures \ref{fig:mmlu_ans_switch_qwen8b}-\ref{fig:tool_choice_tokens_afters_qwen8B} shows distribution of metrics on Qwen3-8B.

\subsubsection{MCQ}

\begin{figure}[H]
    \centering
    \resizebox{\linewidth}{!}{%
        \input{figures/Qwen3-8B/mmlu_pro_bootstrap_metrics_transient}
    }
    \caption{Distribution of the mean of transient flips for raw and denoised outputs in Qwen3-8B, evaluated on the MCQ Questions.}
    \label{fig:mmlu_taf_qwen8b}
\end{figure}

\begin{figure}[H]
    \centering
    \resizebox{\linewidth}{!}{%
        \input{figures/Qwen3-8B/mmlu_pro_bootstrap_metrics_tokens_after}
    }
    \caption{Distribution of the mean of tokens after final answer switch for raw and denoised outputs in Qwen3-8B, evaluated on the MCQ Questions.}
    \label{fig:mmlu_tokens_after_qwen8b}
\end{figure}

\subsubsection{Numeric}

\begin{figure}[H]
    \centering
    \resizebox{\linewidth}{!}{%
        \input{figures/Qwen3-8B/numeric_bootstrap_metrics_answer_switches}
    }
    \caption{Distribution of the mean of answer switches for raw and denoised outputs in Qwen3-8b, evaluated on the Numeric Answers.}
    \label{fig:numeric_ans_switch_qwen8b}
\end{figure}

\begin{figure}[H]
    \centering
    \resizebox{\linewidth}{!}{%
        \input{figures/Qwen3-8B/numeric_bootstrap_metrics_transient}
    }
    \caption{Distribution of the mean of transient flips for raw and denoised outputs in Qwen3-8b, evaluated on the Numeric Answers.}
    \label{fig:numeric_taf_qwen8b}
\end{figure}

\begin{figure}[H]
    \centering
    \resizebox{\linewidth}{!}{%
        \input{figures/Qwen3-8B/numeric_bootstrap_metrics_tokens_after}
    }
    \caption{Distribution of the mean of tokens after final answer switch for raw and denoised outputs in Qwen3-8b, evaluated on the Numeric Answers.}
    \label{fig:numeric_token_after_qwen8b}
\end{figure}

\subsubsection{Search}

\begin{figure}[H]
    \centering
    \resizebox{\linewidth}{!}{%
        \input{figures/Qwen3-8B/search_bootstrap_metrics_answer_switches}
    }
    \caption{Distribution of the mean of answer switches for raw and denoised outputs in Qwen3-8B, evaluated on the Search Query.}
    \label{fig:search_answer_switch_qwen8B}
\end{figure}

\begin{figure}[H]
    \centering
    \resizebox{\linewidth}{!}{%
        \input{figures/Qwen3-8B/search_bootstrap_metrics_transient}
    }
    \caption{Distribution of the mean of transient flips for raw and denoised outputs in Qwen3-8B, evaluated on the Search Query.}
    \label{fig:search_taf_qwen8B}
\end{figure}

\begin{figure}[H]
    \centering
    \resizebox{\linewidth}{!}{%
        \input{figures/Qwen3-8B/search_bootstrap_metrics_tokens_after}
    }
    \caption{Distribution of the mean of tokens after final answer switch for raw and denoised outputs in Qwen3-8B, evaluated on the Search Query.}
    \label{fig:search_taf_qwen8B}
\end{figure}

\subsubsection{Tool Selection}
\begin{figure}[H]
    \centering
    \resizebox{\linewidth}{!}{%
        \input{figures/Qwen3-8B/tool_choice_bootstrap_metrics_answer_switches}
    }
    \caption{Distribution of the mean of answer switches for raw and denoised outputs in Qwen3-8B, evaluated on the Tool Selection.}
    \label{fig:tool_choice_ans_switch_qwen8B}
\end{figure}

\begin{figure}[H]
    \centering
    \resizebox{\linewidth}{!}{%
        \input{figures/Qwen3-8B/tool_choice_bootstrap_metrics_transient}
    }
    \caption{Distribution of the mean of transient flips for raw and denoised outputs in Qwen3-8B, evaluated on the Tool Selection.}
    \label{fig:tool_choice_taf_qwen8B}
\end{figure}

\subsection{Qwen/Qwen3-14B}

\begin{figure}[H]
    \centering
    \resizebox{\linewidth}{!}{%
        \input{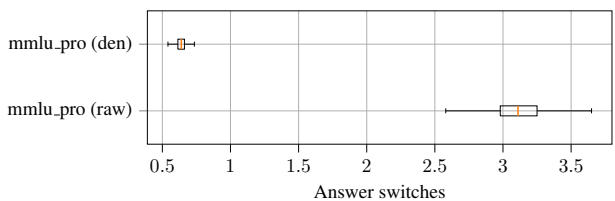}
    }
    \caption{Distribution of the mean of answer switches for raw and denoised outputs in Qwen3-14B, evaluated on the MCQ Questions.}
    \label{fig:mmlu_ans_switch_qwen14b}
\end{figure}

\begin{figure}[H]
    \centering
    \resizebox{\linewidth}{!}{%
        \input{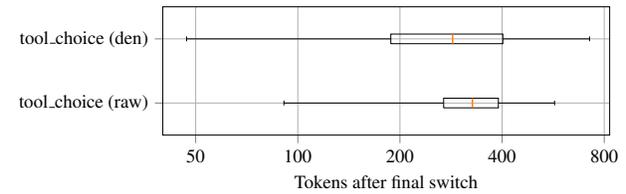}
    }
    \caption{Distribution of the mean of tokens after final answer switch for raw and denoised outputs in Qwen3-14B, evaluated on the Tool Selection.}
    \label{fig:tool_choice_tokens_after_final_qwen14b}
\end{figure}

Figures~\ref{fig:mmlu_ans_switch_qwen14b}-\ref{fig:tool_choice_tokens_after_final_qwen14b} shows the distribution of metrics on different datasets on Qwen3-14B.

\subsubsection{MCQ}

\begin{figure}[H]
    \centering
    \resizebox{\linewidth}{!}{%
        \input{figures/Qwen3-14B/MMLU_TAFs}
    }
    \caption{Distribution of the mean of transient flips for raw and denoised outputs in Qwen3-14B, evaluated on the MCQ Questions.}
    \label{fig:mmlu_tafs_qwen14b}
\end{figure}

\begin{figure}[H]
    \centering
    \resizebox{\linewidth}{!}{%
        \input{figures/Qwen3-14B/MMLU_Tokens_after}
    }
    \caption{Distribution of the mean of tokens after final answer switch for raw and denoised outputs in Qwen3-14B, evaluated on the MCQ Questions.}
    \label{fig:mmlu_tokens_after_qwen14b}
\end{figure}


\subsubsection{Numeric}

\begin{figure}[H]
    \centering
    \resizebox{\linewidth}{!}{%
        \input{figures/Qwen3-14B/Numeric_Answer_switches}
    }
    \caption{Distribution of the mean of answer switches for raw and denoised outputs in Qwen3-14B, evaluated on the Numeric answers.}
    \label{fig:numeric_ans_switch_qwen14b}
\end{figure}

\begin{figure}[H]
    \centering
    \resizebox{\linewidth}{!}{%
        \input{figures/Qwen3-14B/Numeric_TAFs}
    }
    \caption{Distribution of the mean of transient flips for raw and denoised outputs in Qwen3-14B, evaluated on the Numeric answers.}
    \label{fig:numeric_taf_qwen14b}
\end{figure}

\begin{figure}[H]
    \centering
    \resizebox{\linewidth}{!}{%
        \input{figures/Qwen3-14B/Numeric_Tokens_after}
    }
    \caption{Distribution of the mean of tokens after final answer switch for raw and denoised outputs in Qwen3-14B, evaluated on the Numeric answers.}
    \label{fig:numeric_tokens_after_qwen14b}
\end{figure}

\subsubsection{Search}
\begin{figure}[H]
    \centering
    \resizebox{\linewidth}{!}{%
        \input{figures/Qwen3-14B/Search_answer_switches}
    }
    \caption{Distribution of the mean of answer switches for raw and denoised outputs in Qwen3-14B, evaluated on the Search query.}
    \label{fig:search_ans_switch_qwen14b}
\end{figure}

\begin{figure}[H]
    \centering
    \resizebox{\linewidth}{!}{%
        \input{figures/Qwen3-14B/Search_TAFs}
    }
    \caption{Distribution of the mean of transient flips for raw and denoised outputs in Qwen3-14B, evaluated on the Search query.}
    \label{fig:search_taf_qwen14b}
\end{figure}

\begin{figure}[H]
    \centering
    \resizebox{\linewidth}{!}{%
        \input{figures/Qwen3-14B/Search_Tokens_After}
    }
    \caption{Distribution of the mean of tokens after final answer switch for raw and denoised outputs in Qwen3-14B, evaluated on the Search query.}
    \label{fig:search_tokens_after_qwen14b}
\end{figure}

\subsubsection{Tool Selection}

\begin{figure}[H]
    \centering
    \resizebox{\linewidth}{!}{%
        \input{figures/Qwen3-14B/Tool_choice_Answer_switches}
    }
    \caption{Distribution of the mean of answer switches for raw and denoised outputs in Qwen3-14B, evaluated on the Tool Selection.}
    \label{fig:tool_choice_answer_switch_qwen14b}
\end{figure}

\begin{figure}[H]
    \centering
    \resizebox{\linewidth}{!}{%
        \input{figures/Qwen3-14B/Tool_choice_TAFs}
    }
    \caption{Distribution of the mean of transient flips for raw and denoised outputs in Qwen3-14B, evaluated on the Tool Selection.}
    \label{fig:tool_choice_taf_qwen14b}
\end{figure}

\subsection{Qwen3-30B-A3B}
\begin{figure}[H]
    \centering
    \resizebox{\linewidth}{!}{%
        \input{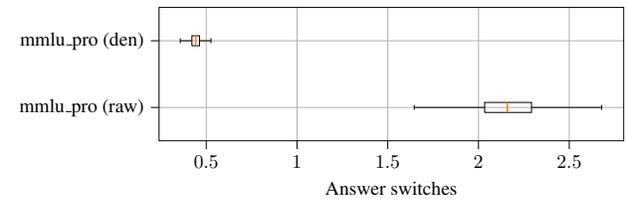}
    }
    \caption{Distribution of the mean of answer switches for raw and denoised outputs in Qwen3-30B-A3B, evaluated on the MCQ Questions.}
    \label{fig:mmlu_ans_switch_qwen30b}
\end{figure}

\begin{figure}[H]
    \centering
    \resizebox{\linewidth}{!}{%
        \input{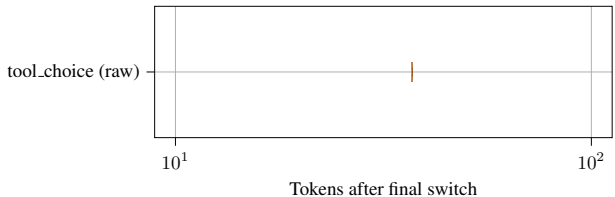}
    }
    \caption{Distribution of the mean of tokens after final answer switch for raw and denoised outputs in Qwen3-30B-A3B, evaluated on the Tool Selection.}
    \label{fig:tool_choice_tokens_afters_qwen30b}
\end{figure}

Figures \ref{fig:mmlu_ans_switch_qwen30b}-\ref{fig:tool_choice_tokens_afters_qwen30b} show bootstrap experiments performed on different datasets.

\subsubsection{MCQ}

\begin{figure}[H]
    \centering
    \resizebox{\linewidth}{!}{%
        \input{figures/Qwen3-30B/mmlu_pro_bootstrap_metrics_transient}
    }
    \caption{Distribution of the mean of transient flips for raw and denoised outputs in Qwen3-30B-A3B, evaluated on the MCQ Questions.}
    \label{fig:mmlu_taf_qwen30b}
\end{figure}

\begin{figure}[H]
    \centering
    \resizebox{\linewidth}{!}{%
        \input{figures/Qwen3-30B/mmlu_pro_bootstrap_metrics_tokens_after}
    }
    \caption{Distribution of the mean of tokens after final answer switch for raw and denoised outputs in Qwen3-30B-A3B, evaluated on the MCQ Questions.}
    \label{fig:mmlu_tokens_after_qwen30b}
\end{figure}

\subsubsection{Numeric}

\begin{figure}[H]
    \centering
    \resizebox{\linewidth}{!}{%
        \input{figures/Qwen3-30B/numeric_bootstrap_metrics_answer_switches}
    }
    \caption{Distribution of the mean of answer switches for raw and denoised outputs in Qwen3-30B-A3B, evaluated on the Numeric Answers.}
    \label{fig:numeric_ans_switch_qwen30b}
\end{figure}

\begin{figure}[H]
    \centering
    \resizebox{\linewidth}{!}{%
        \input{figures/Qwen3-30B/numeric_bootstrap_metrics_transient}
    }
    \caption{Distribution of the mean of transient flips for raw and denoised outputs in Qwen3-30B-A3B, evaluated on the Numeric Answers.}
    \label{fig:numeric_taf_qwen30b}
\end{figure}

\begin{figure}[H]
    \centering
    \resizebox{\linewidth}{!}{%
        \input{figures/Qwen3-30B/numeric_bootstrap_metrics_tokens_after}
    }
    \caption{Distribution of the mean of tokens after final answer switch for raw and denoised outputs in Qwen3-30B-A3B, evaluated on the Numeric Answers.}
    \label{fig:numeric_token_after_qwen30b}
\end{figure}

\subsubsection{Search}

\begin{figure}[H]
    \centering
    \resizebox{\linewidth}{!}{%
        \input{figures/Qwen3-30B/search_bootstrap_metrics_answer_switches}
    }
    \caption{Distribution of the mean of answer switches for raw and denoised outputs in Qwen3-30B-A3B, evaluated on the Search Query.}
    \label{fig:search_answer_switch_qwen30b}
\end{figure}

\begin{figure}[H]
    \centering
    \resizebox{\linewidth}{!}{%
        \input{figures/Qwen3-30B/search_bootstrap_metrics_transient}
    }
    \caption{Distribution of the mean of transient flips for raw and denoised outputs in Qwen3-30B-A3B, evaluated on the Search Query.}
    \label{fig:search_taf_qwen30b}
\end{figure}

\begin{figure}[H]
    \centering
    \resizebox{\linewidth}{!}{%
        \input{figures/Qwen3-30B/search_bootstrap_metrics_tokens_after}
    }
    \caption{Distribution of the mean of tokens after final answer switch for raw and denoised outputs in Qwen3-30B-A3B, evaluated on the Search Query.}
    \label{fig:search_taf_qwen30b}
\end{figure}

\subsubsection{Tool Selection}
\begin{figure}[H]
    \centering
    \resizebox{\linewidth}{!}{%
        \input{figures/Qwen3-30B/tool_choice_bootstrap_metrics_answer_switches}
    }
    \caption{Distribution of the mean of answer switches for raw and denoised outputs in Qwen3-30B-A3B, evaluated on the Tool Selection.}
    \label{fig:tool_choice_ans_switch_qwen30b}
\end{figure}

\begin{figure}[H]
    \centering
    \resizebox{\linewidth}{!}{%
        \input{figures/Qwen3-30B/tool_choice_bootstrap_metrics_transient}
    }
    \caption{Distribution of the mean of transient flips for raw and denoised outputs in Qwen3-30B-A3B, evaluated on the Tool Selection.}
    \label{fig:tool_choice_taf_qwen30b}
\end{figure}

\section{Training Details}
\label{sec:training_details}

\paragraph{Positive and negative training examples.}
Probe training examples are constructed by sampling intermediate reasoning steps from each trace rather than using all steps. Positive examples correspond to reasoning states at which computation could be halted without changing the final answer, whereas negative examples correspond to states where additional reasoning is still required.

For each example, we sample a fixed number \(S\) of reasoning steps from the trace. For a sampled step \(i\), we extract the hidden state of the final generated token at that step from all transformer layers, denoted \(h_i^l\) for layer $l$. Steps satisfying \(i > t^\star\) are assigned a positive label, while steps satisfying \(i \le t^\star\) are assigned a negative label. For examples with \(t^\star = -1\), all sampled steps are treated as positive. The value of \(S\) is treated as a training hyperparameter. 

Because many examples exhibit no answer switch (i.e., \(t^\star = -1\)), positive examples are more frequent than negative examples. To mitigate this imbalance, we oversample negative examples 
so that the two classes are approximately balanced within each minibatch.

\paragraph{Optimization.}
We train the probe using the Adam optimizer. Learning rates are selected from the candidate set
\(\{1\times10^{-4}, 5\times10^{-4}, 1\times10^{-3}, 2\times10^{-3}, 5\times10^{-3}, 1\times10^{-2}, 2\times10^{-2}, 5\times10^{-2}, 1\times10^{-1}\}\).
After fixing the best learning rate, we select the probe layer by choosing the layer whose probe achieves the highest AP on validation.
The probe parameters from the best (learning rate, layer) configuration are used for all reported results for a particular model and dataset.

\section{Experiments on other datasets}
\label{app:aux-datasets}

We present metrics as defined in Section~\ref{sec:answer-trajectory-metrics} on more datasets in Table~\ref{tab:trajectory-summary-aux}

\begin{table*}[t]
\centering
\small

\caption{Summary of answer trajectory metrics on additional datasets.}\label{tab:trajectory-summary-aux}
\begin{tabular}{l l c c c c}
\toprule
Dataset & Accuracy (\%) &
\shortstack{$\Pr(t^\star=-1)$\\Raw\% / Denoised\%} &
\shortstack{AnswerSwitches\\Raw / Denoised} &
\shortstack{TAFs$(T)$} &
\shortstack{$T_{\mathrm{after}}$\\Raw / Denoised} \\
\midrule

\multirow{1}{*}{Humanity's Last Exam}
& 5.2
& 39.2 / 48.9
& 10.8 / 1.3
& 6.5
& 1186 (49.9\%) / 1765 (67.1\%) \\
\midrule
\multirow{1}{*}{GPQA Diamond}
& 56.1
& 35.5 / 43.8
& 10.9 / 1.5
& 6.6
& 1608 (45.7\%) / 2259 (59.8\%) \\
\midrule
\multirow{1}{*}{AIME26}
& 72.7
& 27.2 / 27.2
& 14.3 / 2.1
& 7.1
& 660 (16.4\%) / 1654 (32.9\%) \\

\bottomrule
\end{tabular}
\end{table*}

\end{document}